# AnomalyExplainerBot: Explainable AI for LLM-based anomaly detection using BERTViz & Captum

Prasasthy Balasubramanian[1], Dumindu Kankanamge[2], Ekaterina Gilman[1], and Mourad Oussalah[2]

[1] Faculty of Information Technology and Electrical Engineering, Center for Ubiquitous Computing, University of Oulu, Oulu, Finland 90570
[2] Faculty of Information Technology and Electrical Engineering, Center for Machine Vision and Signal Analysis, University of Oulu, Oulu, Finland 90570

**Abstract.** Conversational AI and Large Language Models (LLMs) have become powerful tools across domains, including cybersecurity, where they help detect threats early and improve response times. However, challenges such as false positives and complex model management still limit trust. Although Explainable AI (XAI) aims to make AI decisions more transparent, many security analysts remain uncertain about its usefulness. This study presents a framework that detects anomalies and provides high-quality explanations through visual tools BERTViz and Captum, combined with natural language reports based on attention outputs. This reduces manual effort and speeds up remediation. Our comparative analysis showed that RoBERTa offers high accuracy (99.6%) and strong anomaly detection, outperforming Falcon-7B and DeBERTa, as well as exhibiting better flexibility than large-scale Mistral-7B on the HDFS dataset from LogHub. User feedback confirms the chatbot's ease of use and improved understanding of anomalies, demonstrating the ability of the developed framework to strengthen cybersecurity workflows.

## 1 Introduction

In the past decade, advances in natural language processing (NLP) have made interactive chatbots valuable tools across various domains. However, their potential in cybersecurity; especially for log analysis; remains largely underutilized. At the same time, cyber-attacks on critical systems have surged, with global losses projected to rise from \$3 trillion in 2015 to \$10.5 trillion [7]. Security analysts play a key role in manually analyzing telemetry data to detect anomalies and isolate real threats, a time-consuming and error-prone task.

To combat these threats, NIST's 2014 Cybersecurity Framework laid out structured policies for identifying, responding to, and recovering from cyber incidents [13]. Building on this foundation, many tools have been developed for Cyber Threat Hunting. However, most automated systems remain black-box in nature, offering little transparency into why a threat was flagged; making it difficult for analysts to trust or act on these insights.

With the emergence of LLMs, there is a growing opportunity to automate threat detection in log analysis. Log file analysis is vital for detecting cyber threats such as unauthorized access and suspicious activity. Traditional rule-based and statistical methods, although widely used, struggle to scale with the complexity of modern systems and depend on predefined features, limiting their ability to detect novel or subtle anomalies [8, 18, 17]. In contrast, LLMs offer a more flexible approach, but their lack of interpretability raises concerns, particularly in high-stakes domains, like cybersecurity. As global standards emphasize responsible and explainable AI, it is crucial that AI-based systems also support human-centered understanding and build trust.

This research addresses this critical gap by proposing *AnomalyExplainer* Bot; a conversational AI framework that not only detects anomalies in log data, but also explains its reasoning through intuitive visualizations (using tools like BERTViz and Captum) and natural, dialogue-based reports. This approach enables faster, more reliable decision making while reducing manual burden and improving analyst confidence in AI-assisted threat detection.

In overall, the key contributions of this research are as follows:

1. *Development of a Conversational AI Framework:* We present a novel conversational AI system that uses advanced LLMs to detect anomalies in log data and explain them interactively through a user-friendly chatbot interface.
2. *Integration of Explainability Tools:* The framework incorporates high-end Explainable AI(XAI) tools such as BERTViz and Captum to analyze and justify LLM's decisions. These visual explanations are seamlessly integrated into the user interface to enhance interpretability.
3. *Conversational Explainability for Enhanced Trust:* The system offers conversational assistance focused on explaining model decisions, increasing trust in LLM outputs. This improves usability, reduces manual effort, and encourages a wider adoption of LLMs in cybersecurity workflows.
4. *Comparative Study of LLM Architectures:* We conduct a comparative analysis of encoder-based and decoder-based LLMs to identify models that are efficient and lightweight for real-time anomaly detection. RoBERTa was ultimately selected as the optimal choice due to its high accuracy and practical performance.
5. *Practical Implementation, User Study, and Insights:* We demonstrate the effective deployment of the proposed framework in a cybersecurity context, supported by a user study that highlights its usability, interpretability, and impact on analyst trust. The study also discusses infrastructure challenges and considerations for real-world adoption.

## 2 Related Work

Log file analysis is a critical component of cybersecurity, enabling the detection and mitigation of threats originating from various sources. This section reviews key literature to contextualize our study and identify current challenges. Specifically, we examine research in three areas: anomaly detection methods, applications of XAI, and the integration of conversational agents in cybersecurity. Numerous studies have highlighted the significance of identifying anomalies in log data, which often signal unauthorized access or potential breaches [18, 16]. Consequently, log analysis has become a prominent focus in cybersecurity research [1]. To overcome these limitations, ML techniques have been widely adopted. However, despite their success in many domains, the application of LLMs for log anomaly detection remains relatively rare. Recent works have begun to explore this potential. For instance, [5] introduced a conversational anomaly detection framework that uses GPT-3.5 for dialogue generation and ALBERT for anomaly detection, achieving over 99% accuracy. Similarly, [9] proposed LLMeLog, which uses in-context learning to enrich log events and fine-tunes a pre-trained BERT model. These enriched embeddings are then passed to a transformer-based anomaly detector, achieving a F1-score exceeding 90% with only 10% labeled data.

Despite these advancements, industry adoption of LLM-based solutions for cybersecurity remains limited. One of the primary concerns is the “black-box” nature of these models, which makes it difficult for analysts to understand or trust their decisions. Although some studies, such as [2], attempt to bridge this gap by combining GPT-3.5 with a Random Forest classifier and incorporating explainability tools like LIME and SHAP, these methods are often not well-suited for complex transformer-based architectures. LIME and SHAP provide featurelevel explanations but fall short in capturing the nuanced internal mechanisms and attention flows characteristic of LLM.

While conversational agents have gained attention in cybersecurity applications [4, 2], most existing implementations are limited in scope. They often lack integration with sophisticated backend models and do not provide robust explainability components. Moreover, the use of open-source LLMs for dialogue generation in cybersecurity remains an understudied area, despite the potential for reducing manual effort and enhancing analyst productivity.

Another important gap is the under-utilization of advanced interpretability tools specifically designed for transformer-based models, such as BERTViz and Captum. These tools provide deep insights into attention mechanisms and model attributions, which can be vital to understanding why a particular decision was made. However, some cybersecurity studies have leveraged these capabilities.

The importance of explainability is further underscored by recent reviews such as [15], which offers a comprehensive taxonomy of XAI methods relevant to cybersecurity. Their work emphasizes the need for human-understandable, reliable AI explanations to build trust, ensure accountability, and enable intelligent decision-making in security-critical environments.

In summary, our review highlights three key gaps in current research: (1) the limited application of LLMs for anomaly detection in log analysis; (2) the lack of conversational interfaces powered by open source LLMs in cybersecurity; and (3) the insufficient use of advanced explainability tools to interpret LLM behavior. This study addresses these challenges by introducing the AnomalyExplainer Bot-a conversational AI framework that combines efficient anomaly detection with transparent, interactive explanations using BERTViz and Captum. Our approach is designed to enhance trust, reduce manual workload, and support real-world adoption of LLMs in cybersecurity operations.

# 3 Methodology

## 3.1 Log Anomaly Detection

Our chatbot's core function is anomaly detection in system logs. It analyzes user-submitted logs to identify deviations from expected behavior. The final implementation uses a fine-tuned encoder-based RoBERTa (roberta-base) model trained on a subset of the annotated LogHub dataset [20]. The pipeline involves preprocessing and normalizing raw logs, tokenization and feeding tokens to the RoBERTa encoder to generate contextual embeddings. A classification head predicts whether each log entry is anomalous. Further post-processing enables formatting and visualization of the results. The output is then translated into a human-readable response using the chatbot dialogue generation model (GPT-4o). Figure 1 presents the sequence flow of the AnomalyExplainer bot from log file upload to response delivery. The user uploads a data file via the chatbot interface, which stores the file and initiates anomaly detection. The anomaly detection module processes each segment of the file, identifying any irregularities. Based on the result of the detection, the response generator prepares either a standard (no anomaly detected) or an anomaly-specific response. If anomalies are found, users may request explanations, triggering the explainability module to retrieve relevant context and generate human-interpretable insights. Finally, the chatbot delivers responses and logs the entire interaction for traceability in a MongoDB database.

## 3.2 Explainable AI components

To improve LLM explainability, we used BERTViz and Captum. The system provides two types of output to support the user's understanding of the LLM's decision-making process. A conversational report generated from these tools and interactive visualizations accessible via a separate tab in the bot interface through an iFrame. Users can select specific log lines from a dropdown to explore detailed explainability.

BERTViz [19] is an open-source tool for visualizing self-attention in BERT-based models at attention-head, model, and neuron levels. Our study focuses on attention-head and model views. When users upload a validated log file and click "Find Anomalies," a POST request to the Django server processes the file, assigns a session ID, and triggers the explainability engine. BERTViz then generates head-view and model-view visualizations for each log entry, storing attention details in MongoDB. Users access these via the "Visualizations" button in the interface. Captum, developed by Facebook AI Research [11], provides interpretability tools for PyTorch models, offering gradient- and perturbation-based attribution methods to assess feature, neuron, and layer importance. Captum Insights extends this with interactive visualizations for sample-based model debugging. We generate a conversational report by applying a unified analysis algorithm (Algorithm 1) to outputs from BERTViz and Captum [6]. The algorithm first computes per-token attention scores by averaging attention received across all heads and layers, normalizing, and extracting the top-$k$ tokens. It then identifies the most focused attention heads by calculating entropy of their attention distributions, selecting the heads with lowest entropy score. At the layer level, it averages the inverse entropy of the heads to find the most focused layers. The algorithm also checks for attention bias toward special tokens (e.g., [CLS], [SEP], <s>) and issues warnings if the biases exceed a given threshold. Finally, it compiles these insights, consisting of top tokens, focused heads and layers, and bias warnings, into a structured report supporting model interpretability.

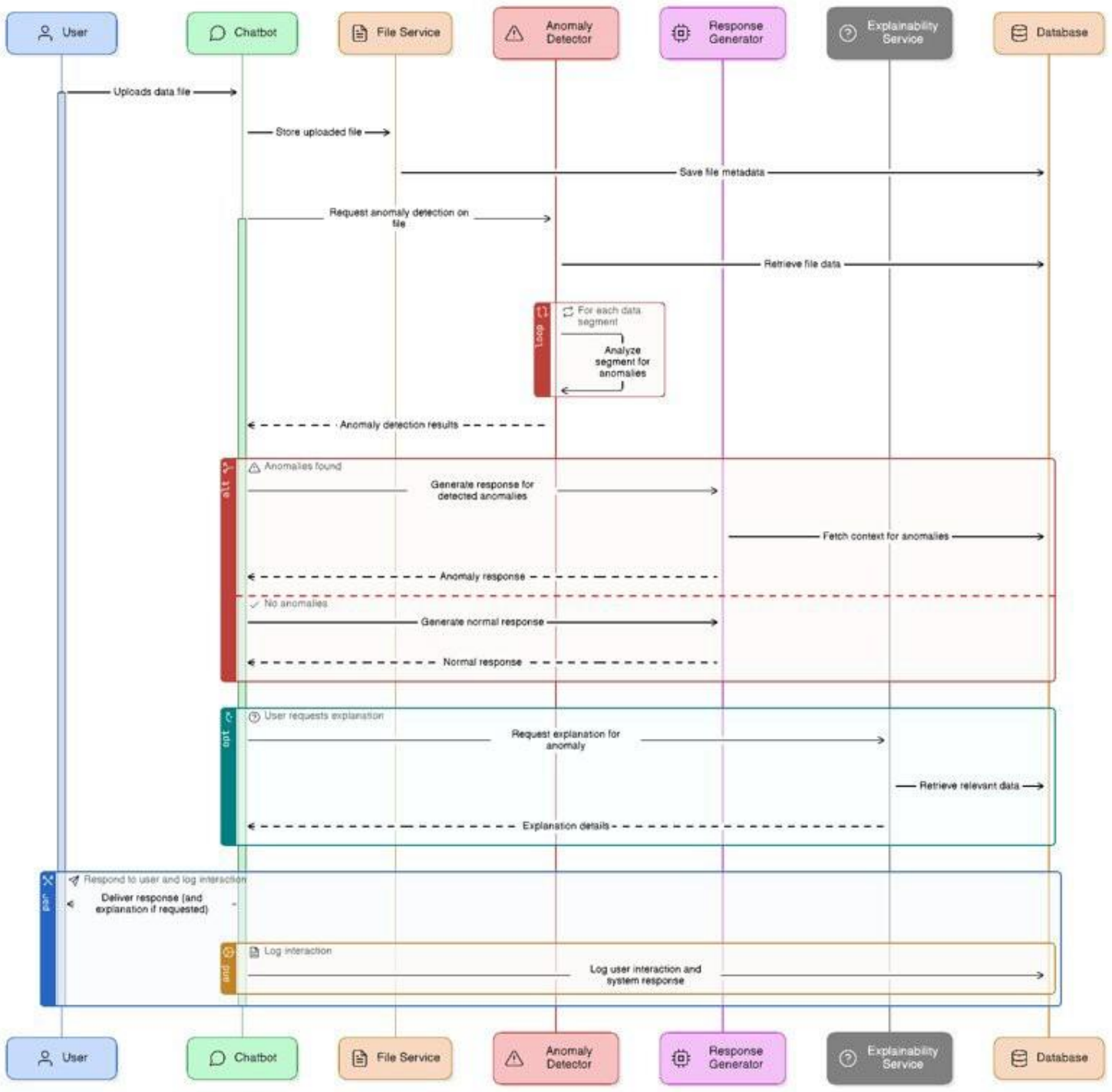

Fig. 1: Sequence diagram of AnomalyExplainer bot

In summary, the system uses post-hoc, attention-based explainability, analyzing transformer attention patterns with entropy-based focus measures and bias detection. It also integrates feature attribution via Captum. Explanations combine textual summaries and interactive visualizations (BERTViz, Captum) in separate UI tabs, providing clear, human-centered insights into anomaly detection.

### 3.3 System Architecture and Technical Stack

Figure 2 shows the architecture of the explainable AnomalyExplainer bot. Users upload log files and request analysis through the chatbot frontend, which sends input to a LangChain-powered backend combining an OpenAI dialogue model, anomaly detection, and a database. The anomaly detection module analyzes logs, detects anomalies, and generates explanations using BERTViz (attention visualization) and Captum (feature attribution). Explanation data is

**Algorithm 1:** Unified Attention Analysis

**Data:** attentions, tokens, num_layers, num_heads, seq_len, top_k_tokens, top_k_heads, top_k_layers, special_tokens, bias_threshold
**Result:** Formatted report string with attention insights
Initialize $token_scores \leftarrow \mathbf{0}_{seq_len}$;
Initialize $focused_heads \leftarrow []$;
Initialize $layer_scores \leftarrow []$;
Initialize $bias_info \leftarrow []$;
**for** $layer_idx = 0$ **to** $num_layers - 1$ **do**
  $total_focus \leftarrow 0$;
  **for** $head_idx = 0$ **to** $num_heads - 1$ **do**
    $head_attn \leftarrow attentions[layer_idx][head_idx]$;
    // Accumulate token attention scores averaged over queries (axis=0)
    $token_scores \leftarrow token_scores + \text{mean}(head_attn, axis = 0)$;
    // Calculate entropy for each query distribution in this head
    $entropies \leftarrow -$ � $(head_attn \odot \log(head_attn + 10^{-9}), axis = -1)$;
    $avg_entropy \leftarrow \text{mean}(entropies)$;
    $focused_heads$.append(($layer_idx, head_idx,$ round($avg_entropy,$ 3)));

    // Accumulate inverse entropy to total focus for layer
    $total_focus \leftarrow total_focus + \frac{1}{avg_entropy+10^{-9}}$;
    // Check special token bias **for** $special_token \in special_tokens$ **do**
      **if** $special_token \in tokens$ **then**
        $idx \leftarrow tokens$.index($special_token$);
        $avg_focus \leftarrow \text{mean}(head_attn[:, idx])$;
        **if** $avg_focus > bias_threshold$ **then**
          $bias_msg \leftarrow$ format("Bias detected: Layer layer_idx, head_idx, special_token, avg_focus) bias_info.append(bias_msg) **end**
        **end**
      **end**
    **end**
    $avg_focus \leftarrow total_focus/num_heads$;
    $layer_scores$.append(($layer_idx,$ round($avg_focus,$ 3)));
  **end**
  $token_scores \leftarrow token_scores/(num_layers \times num_heads)$;
  $top_token_indices \leftarrow \text{argsort}(token_scores)[-top_k_tokens:][::-1]$;
  $top_tokens \leftarrow [(tokens[i], \text{round}(token_scores[i], 3))$ for $i \in top_token_indices]$;
  $top_focused_heads \leftarrow \text{sort}(focused_heads, key = \lambda x : x[2])[: top_k_heads]$;
  $top_layers \leftarrow \text{sort}(layer_scores, key = \lambda x : x[1], reverse = True)[: top_k_layers]$;
  Build *report* $\leftarrow$ formatted string including:
  - **Top Attended Tokens:** *top_tokens*
  - **Most Focused Heads (lowest entropy):** *top_focused_heads*
  - **Standout Layers (highest focus scores):** *top_layers*
  - **Special Token Bias Warnings:** *bias_info* (if not empty, else "None")
  **return** *report*;

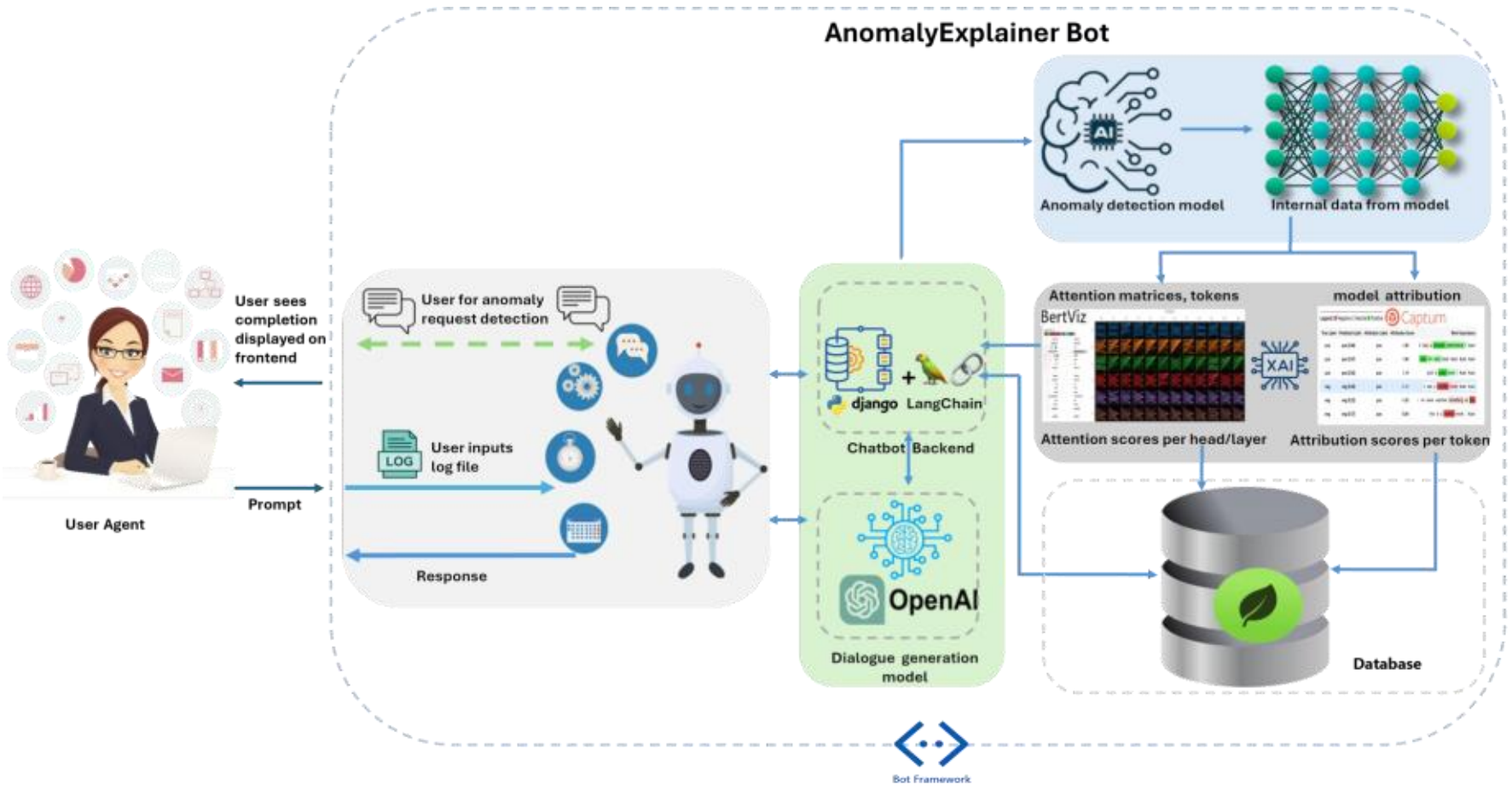

Fig. 2: System architecture for Bot framework

stored for transparency. The chatbot then creates human-readable responses and sends them back to users, closing the feedback loop. The system runs on the Bot Framework, enabling smooth interaction, with visualizations accessible in the UI via an iFrame. The backend uses Django (v5.1.5) with Jinja templates, and the frontend relies on basic HTML, CSS, and JavaScript. Key libraries include Hugging Face Transformers, Torch (with Torchaudio, Torchvision), PEFT, Captum, BERTviz, and BitsAndBytes for ML and chatbot functions. The chatbot runs GPT-4o via the OpenAI API, managed by LangChain and monitored with LangSmith. Visualizations use Matplotlib, Seaborn, Pandas, and IPython. MongoDB (via MongoEngine) stores log and app data, while SQLite manages users in Django. This modular stack enables efficient development, deployment, and monitoring of the anomaly detection chatbot. LangChain helps build complex LLM workflows by handling prompts and external data integration. Detailed tech stack info is in Table 1.

### 3.4 LLM-Based Comparative Study

In this study, we adopted a comparative approach to benchmark transformer-based models, including encoder-focused (RoBERTa, DeBERTa) and decoder-focused (Falcon-7B, Mistral-7B Yarn) architectures. These models were chosen for their widespread usage, accessibility, and suitability for experimentation, with the availability of pre-trained weights being a key factor. Each model was trained on 4,000 samples, validated on 500, and evaluated after 3 epochs using a separate test set of 500 samples. Due to their large size, training and evaluation presented significant computational challenges. Our evaluation focused on accuracy and standard classification metrics. Model implementation and management were handled using the Hugging Face framework, which streamlined access to pre-trained models and supported tasks like text classification, NER, and sentiment analysis. A brief summary of the selected models follows.

Table 1: Tech Stack Overview

| Category | Tool / Library | Purpose |
|---|---|---|
| Core Frameworks and Infrastructure | Python | Primary programming language |
| | Django (5.1.5) | Backend web framework |
| | Django REST Framework (3.15.2) | API development |
| | Gunicorn (23.0.0) | WSGI HTTP server for deployment |
| Machine Learning and Model Inference | PyTorch (2.6.0) | Deep learning framework |
| | Transformers (4.48.2) | Access to pretrained LLMs |
| | Accelerate (1.3.0) | Efficient multi-GPU model execution |
| | Bitsandbytes (0.45.1) | Optimized low-precision computation for LLMs |
| Explainability and Visualization | Captum (0.8.0) | Interpretability for PyTorch models |
| | BERTViz (1.4.0) | Visualization of transformer attention |
| | Matplotlib (3.10.1) | Static data visualization |
| Conversational Intelligence and Observability | LangChain (core==0.3.33, openai==0.3.3) | LLM orchestration and prompt chaining |
| | LangSmith (0.3.4) | Debugging, observability, and evaluation of LLM applications |
| Database and Data Handling | MongoEngine (0.29.1) | ODM for MongoDB |
| | PyMongo (4.11.3) | MongoDB interface |
| | Python-dotenv (1.0.1) | Environment variable management |

**RoBERTa – A Robustly Optimized BERT Pretraining Approach** RoBERTa, developed by Facebook AI [12], improves BERT's training by removing the Next Sentence Prediction (NSP) objective and using larger mini-batches, longer sequences, and more data. It retains the bidirectional encoder architecture and Masked Language Modeling (MLM) objective. RoBERTa uses Byte-Pair Encoding (BPE) and is available in variants such as RoBERTa-base (125M parameters) and RoBERTa-large (355M). It performs strongly on tasks like text classification and QA. We used RoBERTa-base in our study.

**DeBERTa – Decoding-enhanced BERT with Disentangled Attention** DeBERTa by Microsoft Research [10] introduces disentangled attention, separating content and positional information, and an enhanced mask decoder that incorporates absolute positional embeddings. With configurations ranging from 100M to 1.5B parameters, it achieves strong results on benchmarks like MNLI and SQuAD. We utilize the DeBERTa-base (100M) model for its balance of efficiency and accuracy.

**Falcon-7B** Falcon-7B, developed by the Technology Innovation Institute [3], is a 7B-parameter decoder-only transformer optimized for text generation, summarization, and QA. It incorporates sparse attention and mixed-precision training for efficient performance. Trained on large, diverse datasets and using SentencePiece tokenization, Falcon-7B offers strong generalization. We include it in our study for its robust language modeling capabilities.

**Mistral-7B (Yarn)** Mistral-7B (Yarn) [14] is a decoder-only transformer by Mistral AI that supports up to 128k context tokens using the YaRN extension to RoPE. It uses sliding-window and grouped-query attention for efficient long-context processing. Featuring a SentencePiece (BPE) tokenizer and fast inference, it is well-suited for tasks like summarization, code generation, and classification.

We use this model in our classification experiments. More details of the models used are summarized in Table 2.

Table 2: Comparison of Selected Transformer Models

| Model | Details |
|---|---|
| RoBERTa-base [10] | **Developer:** Facebook AI<br>**Architecture:** Encoder-only (BERT-based)<br>**Key Features:**<br>- Removes NSP objective<br>- Trained on longer sequences & larger mini-batches<br>- Uses more training data<br>- MLM objective retained<br>**Tokenizer:** Byte-Pair Encoding (BPE)<br>**Parameters:** ~125M (base), ~355M (large)<br>**Applications:** Text classification, QA, sentiment analysis, etc. |
| DeBERTa-base [10] | **Developer:** Microsoft Research<br>**Architecture:** Encoder-only (BERT-based)<br>**Key Features:**<br>- Disentangled attention (content & position)<br>- Enhanced mask decoder for MLM<br>- Superior performance on MNLI, SQuAD<br>**Tokenizer:** Byte-level BPE<br>**Parameters:** ~100M (base), ~350M (large), up to 1.5B<br>**Applications:** Text classification, QA, sentiment analysis, etc. |
| Falcon-7B [3] | **Developer:** Technology Innovation Institute<br>**Architecture:** Decoder-only (Transformer)<br>**Key Features:**<br>- Sparse attention, mixed precision training<br>- Highly efficient with strong generalization<br>- Optimized for language understanding & generation<br>**Tokenizer:** SentencePiece<br>**Parameters:** 7B<br>**Applications:** Text generation, summarization, QA, etc. |
| Mistral-7B (Yarn) [14] | **Developer:** Mistral AI<br>**Architecture:** Decoder-only (Transformer)<br>**Key Features:**<br>- YaRN: RoPE extended to 128k context<br>- Sliding-window & grouped-query attention<br>- Long-context support, fast inference<br>- Trained on high-quality datasets<br>**Tokenizer:** SentencePiece (BPE)<br>**Parameters:** 7B<br>**Applications:** Summarization, code generation, dialogue, classification, etc. |

## 3.5 Performance Metrics

When evaluating classification models, key performance metrics-Accuracy, Precision, Recall, and F1-score, are derived from True Positives (TP), True Negatives (TN), False Positives (FP), and False Negatives (FN). As illustrated in Table 3, Accuracy reflects the overall proportion of correct predictions. While Precision measures the proportion of true positives among predicted positives, indicating the model's ability to minimize false positives. Recall (or sensitivity) captures how well the model identifies actual positives. F1-score, the harmonic mean of Precision and Recall, balances the trade-off between false positives and false negatives.

## 3.6 Experimental Settings

We used data from LogHub [20], specifically the HDFS subset containing over 11 million labeled log messages from Amazon EC2, intended for anomaly detection. Logs include block-level operations (e.g., allocation, writing, replication) and are labeled using sequence-based methods. We reshuffled the data and extracted normal and anomalous records to train our classifier. Training and evaluation were conducted on RunPod.io using a PyTorch 2.1 environment with an RTX 3090 GPU (24 GB VRAM), 125 GB RAM, 16 vCPUs, and 200 GB storage. Models such as RoBERTa, DeBERTa, Falcon 7B, and Yarn-Mistral 7B were trained on 4,000 samples, validated on 500, tested on 500, and evaluated after 3 epochs.

Table 3: Classification performance metrics based on TP, TN, FP, and FN.

| Metric | Formula |
|---|---|
| Accuracy | $(TP+TN)/(TP+TN+FP+FN)$ |
| Precision | $TP/(TP+FP)$ |
| Recall (Sensitivity) | $TP/(TP+FN)$ |
| F1-score | $2TP/(2TP+FP+FN)$ |

The application was deployed on CSC's cPouta cloud platform using a virtual machine with the standard.xlarge flavor (6 vCPUs, 15 GiB RAM, 80 GB disk). This configuration offers sufficient resources for inference ML services and interactive visualizations, expected to provide a scalable and reliable environment for hosting AI applications.

# 4 Results

## 4.1 AnomalyExplainer Chatbot Interface and dialogues

Figure 3 shows the main interface of the AnomalyExplainer chatbot. The top-left panel displays the initial anomaly detection output, detailing the event, severity, possible causes, and recommended actions. The right panel presents an explainability report generated using unified analysis algorithm, highlighting key tokens, dependencies, and contextual patterns. Feedback form shown at the bottom gathers user demographics and qualitative input for iterative improvement. Figure 4 illustrates the Explainable AI tabs, featuring interactive BERTViz attention visualizations and Captum-based attributions. Users can select specific log entries via a dropdown to explore attention layers, token relationships, and word-level importance.

The bot's anomaly detection dialogue responses show moderate to high linguistic complexity, with readability metrics like the Flesch-Kincaid Grade(10.17) and Gunning Fog Index(12.18) indicating reading levels around 10th to 12th grade. This makes them suitable for technical users like security analysts and system admins but possibly too complex for general audiences. The explainability conversational reports have even higher complexity, with metrics such as the Flesch Reading Ease(38.08) and SMOG Index(13.9) showing late high school to early college levels (grades 12–14), fitting expert users but likely challenging for non-technical readers.

## 4.2 Comparative analysis results

Table 4 and Figure 5 compare the performance of four transformer-based models on a binary classification task (normal vs. anomaly). Mistral-7B (Yarn) achieved perfect results across all metrics, correctly identifying both normal and anomalous cases. RoBERTa also performed very well, with high accuracy and

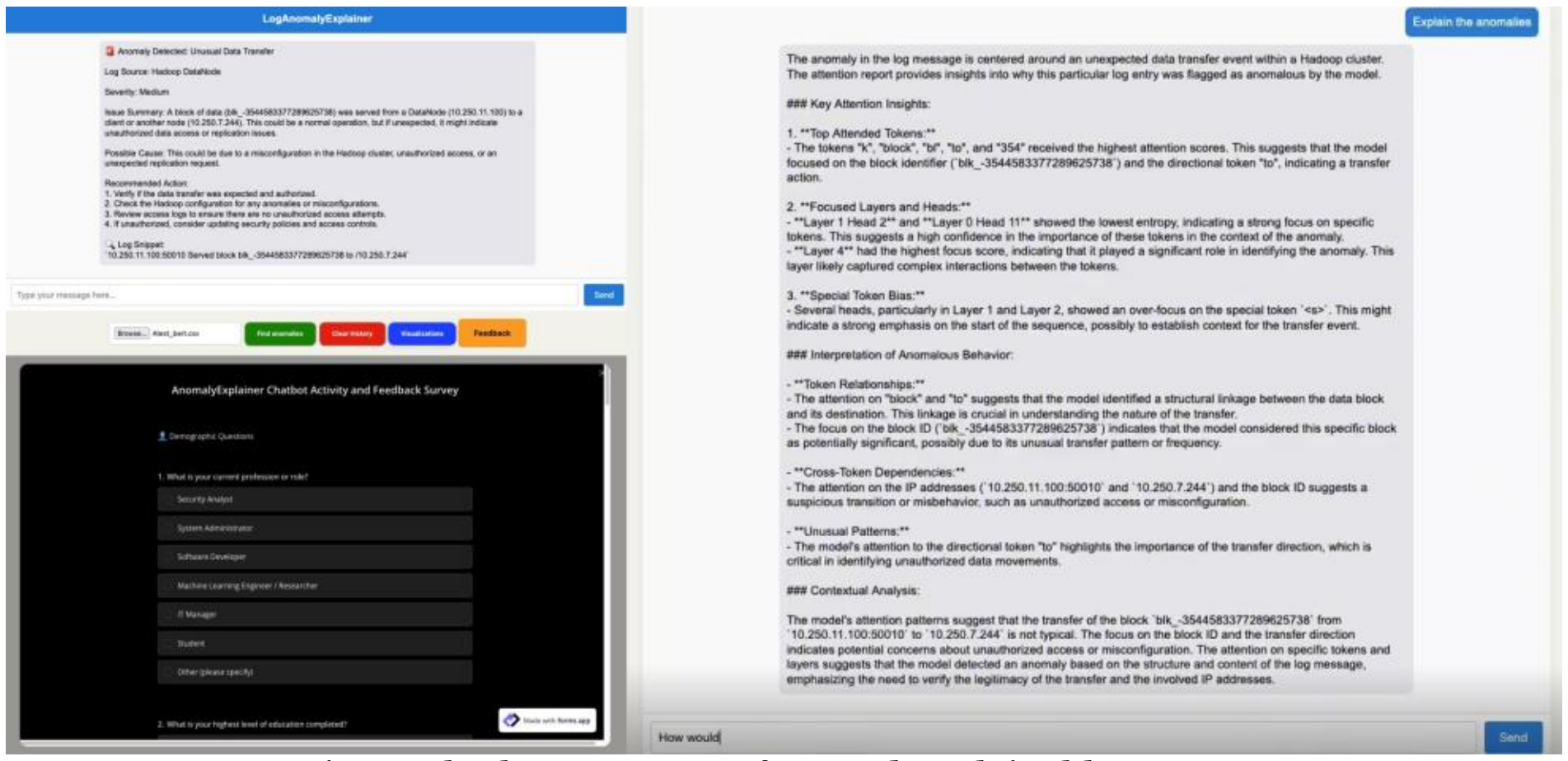

Fig. 3: Chatbot User Interface and explainable report

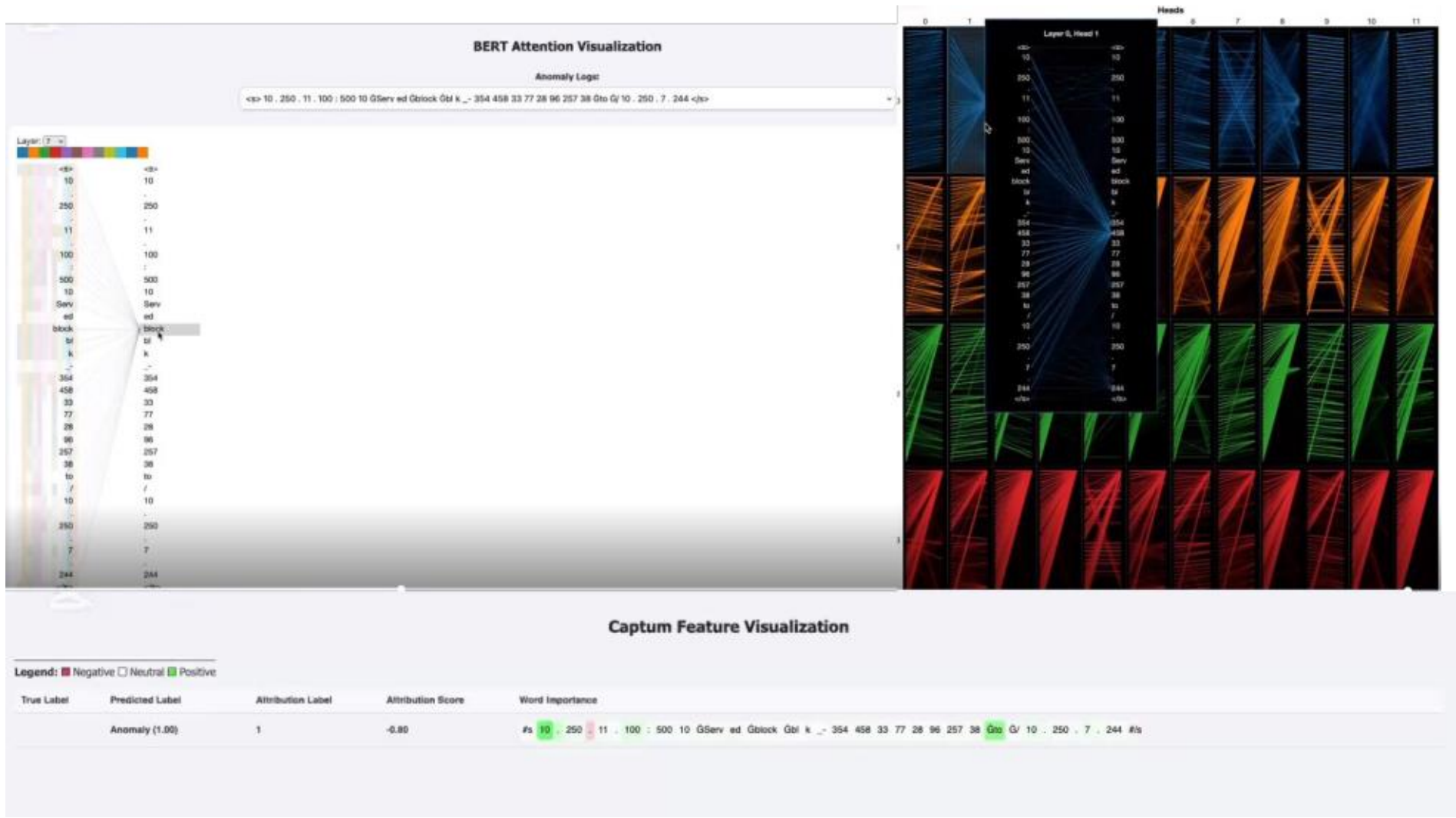

Fig. 4: Screenshots of Explainable AI interface

F1-score, though its anomaly recall dropped slightly to 0.91. DeBERTa showed strong accuracy (0.9840) and good performance on normal data, but its lower anomaly recall (0.65) reduced its F1-score for anomalies (0.79) and macro-average F1 to 0.89. Falcon-7B had high accuracy (0.9540) and perfect recall for normal data but failed to detect any anomalies, scoring 0.00 F1 for the anomaly class and a low macro-average F1 of 0.49. These results highlight the need to evaluate models not just on accuracy but also on how well they detect rare events like anomalies.

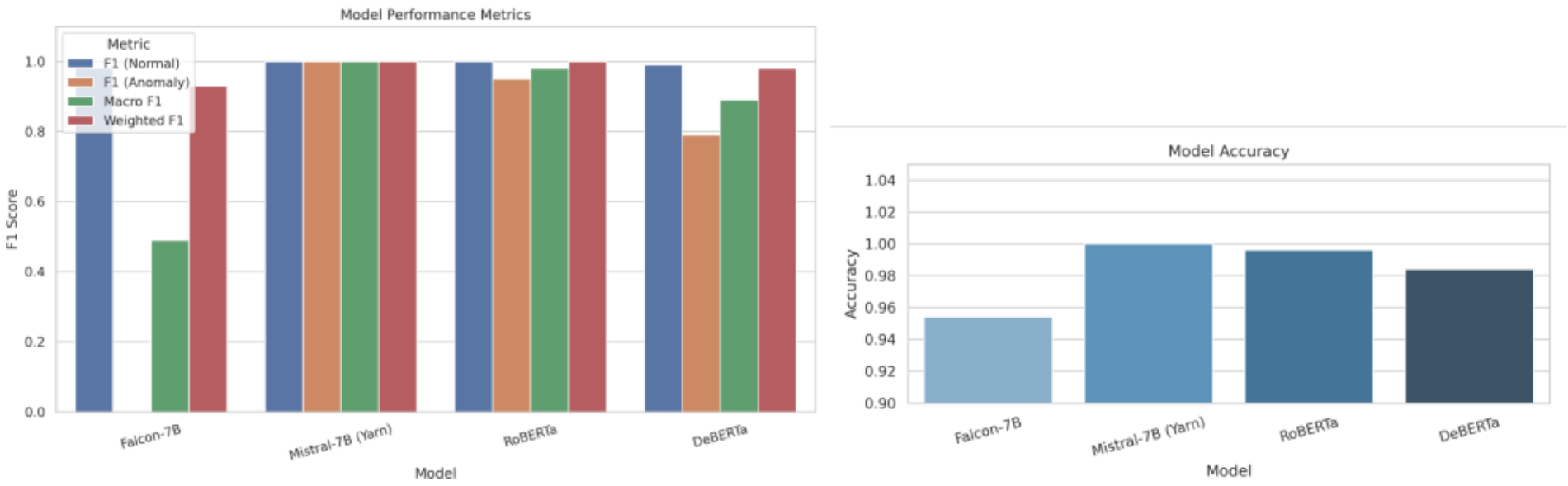

Fig. 5: Model performance metrics comparison

Table 4: Performance comparison of models (Normal and Anomaly classes).

(a) Performance on Normal class

| Model | Accuracy | Precision (N) | Recall (N) | F1 (N) |
|---|---|---|---|---|
| Falcon-7B | 0.9540 | 0.95 | 1.00 | 0.98 |
| Mistral-7B (Yarn) | 1.0000 | 1.00 | 1.00 | 1.00 |
| RoBERTa | 0.9960 | 1.00 | 1.00 | 1.00 |
| DeBERTa | 0.9840 | 0.98 | 1.00 | 0.99 |

(b) Performance on Anomaly class and overall

| Model | Precision (A) | Recall (A) | F1 (A) | Macro F1 | Weighted F1 |
|---|---|---|---|---|---|
| Falcon-7B | 0.00 | 0.00 | 0.00 | 0.49 | 0.93 |
| Mistral-7B (Yarn) | 1.00 | 1.00 | 1.00 | 1.00 | 1.00 |
| RoBERTa | 1.00 | 0.91 | 0.95 | 0.98 | 1.00 |
| DeBERTa | 1.00 | 0.65 | 0.79 | 0.89 | 0.98 |

# 5 User Study and Analysis

## 5.1 Study Design and Protocol

We conducted a structured user study with 13 participants to assess the usability, explainability, and trustworthiness of our anomaly detection system. Participants interacted with a chatbot-based log analysis interface, uploaded log files, triggered anomaly detection, viewed attention-based explanation reports, and explored visualizations (BERTViz, Captum).Feedback was collected through 12 targeted questions to assess usability, interpretability, trust, and overall user experience and 2 demographic items, embedded within the UI for accessibility.

The system generates unique session IDs per analysis, stores predictions and related data in a MongoDB backend, and presents results via a GPT-4o-powered conversational interface.Participants completed a questionnaire with multiple-choice items on usability, trust, explanations, and visuals, plus two open-ended questions for feedback.

**Participant Demographics** Participants came from diverse technical roles (Table 5), with most holding graduate degrees (9 Master's, 2 PhDs), reflecting a highly technical user base.

## 5.2 Results and Insights

**Usability and Interaction** The system received positive usability feedback:

– 10 out of 13 users rated chatbot interaction as *easy* or *very easy*.

Table 5: Participant professional backgrounds.

| Profession | Count | Profession | Count |
|---|---|---|---|
| Software Dev. | 3 | ML Eng. / Researcher | 3 |
| Student | 2 | Sys Admin | 2 |
| Security Analyst | 1 | Site Reliability Eng. | 1 |
| Teacher (Other) | 1 | | |

– 9 users found uploading and analyzing logs *smooth and intuitive* or experienced only minor issues.

*"The chatbot interface is simple and intuitive. I was able to understand my logs better with the explanations provided."* – Participant

**Trust and Explanations Quality**

– 7 participants indicated they *somewhat trust* the anomaly predictions.
– Only 2 participant reported full trust; one expressed low trust.
– The chatbot's explanations were rated *helpful* or *very helpful* by a majority.
– Suggested actions and anomaly reasons were identified as the most useful elements.

**Visualization and XAI Tools** Feedback revealed challenges with visualization tools:

– 7 participants find it not useful or didn't use the visualization tools.
– Only 4 participants rated them as *very useful*.
– Common issues included timeouts, HTML artifacts in the output, and lack of interpretability.

*"I couldn't understand what the visualization was showing. It timed out or returned HTML tags. More guidance is needed."* – Participant

While some users valued the attention-based reports, many found them too technical or insufficiently explained for non-ML audiences.

**Key Themes from Open Feedback** Open-ended feedback (Table 6) revealed key improvement areas: UI/UX clarity, system responsiveness, filtering features, and better visualization guidance.

Table 6: Qualitative themes identified from open-ended user feedback.

| Theme | Comment Frequency | Example Quote |
|---|---|---|
| Visualization Unreliable | High | "Captum timed out", "BERTViz didn't show anything", "No error shown" |
| Chatbot Explanations Helpful | Medium | "Chatbot's recommended actions were very helpful" |
| UI/UX Improvements Needed | High | "Unclear buttons", "Timeouts with no feedback" |
| System Lag & Session Stability | Medium | "System stops functioning after a few interactions" |
| Desire for Filtering/Context | Medium | "Should filter by node address", "Show filename with analysis" |

**Summary of User Study** The study (Table 7) shows users found the chatbot interface highly usable and valued the AI-generated explanations. However, tools like BERTViz and Captum were often underused or non-functional, indicating a key area for improvement. Technical trust was moderate, suggesting the need for clearer, more structured reports. These findings will guide the next system iteration.

Figure 6 summarizes participant responses. Most users (8) found the chatbot easy to use, and 9 found the upload and analysis process smooth. Trust was moderate, with 7 partially trusting the anomaly predictions and only one fully trusting. Visual explanation tools were less useful, with 7 participants not using or finding them unhelpful. Users were mostly technically proficient, matching the target audience, though results should be interpreted cautiously.

Table 7: Summary of key user study findings across demographic, usability, and trust dimensions.

| **Question/Area** | **Most Common Response / Observation** | **Secondary Insight** |
|---|---|---|
| Profession | Software Developer (3) | Student / System Administrator |
| Education | Master's Degree (9) | PhD (2) |
| Ease of Chatbot Interaction | Easy (8) | Very Easy (2), Neutral (3) |
| Upload & Analysis | Smooth and intuitive (7) | Minor issues (4), 1 confused |
| Trust in Predictions | Somewhat trusted (7) | Full trust (2), Low trust (1) |
| Visualization Tools | Not useful / Not used (7) | Very useful (4), unclear output noted |
| Explainability | Most users found chatbot's NL explanations helpful | Suggested actions and reasons were most valued |
| Technical Challenges | Timeouts, UI unresponsiveness | HTML artifacts, need for context |

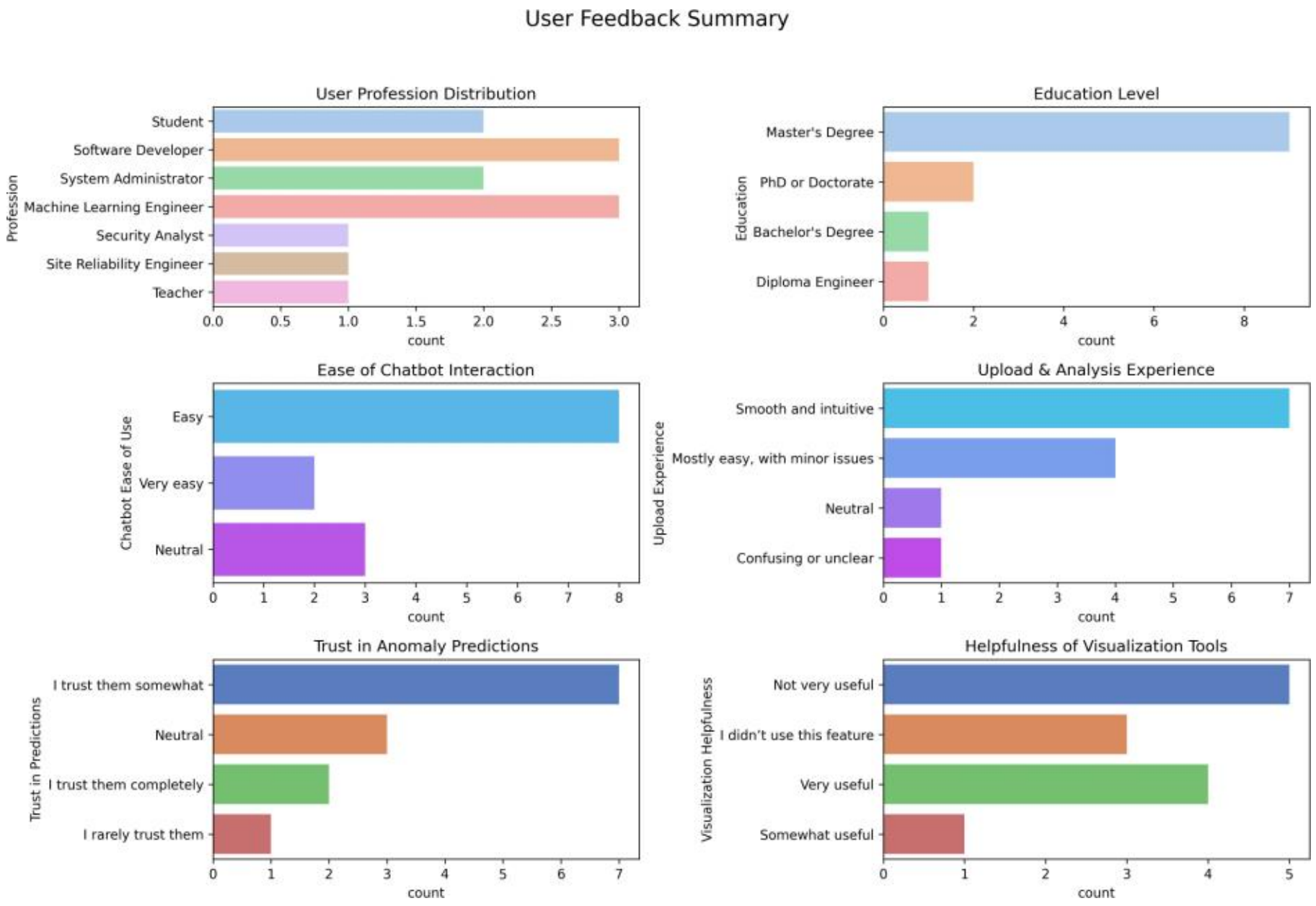

Fig. 6: Overview of user feedback across profession, education, usability, trust, and feature usefulness.

# 6 Discussion and Future Work

This research aims to create a user-friendly interface for security analysts to quickly analyze log files, detect anomalies, and build trust in AI through explainable insights. The system uses post-hoc, attention-based explanations with feature attribution and interactive visualizations to provide clear, human-centered insights. RoBERTa was chosen for the final framework due to its strong balance of high accuracy (99.6%) and reliable anomaly detection. Although Mistral-7B performed slightly better in recall, it was too large and complex to integrate efficiently. User feedback confirmed that the chatbot interface was intuitive and its explanations helped users better understand anomalies, improving trust and usability. However, the visualization tab sometimes loads slowly or fails due to heavy backend computations by BERTViz and Captum, especially with large logs or complex models. Additional delays are caused by iFrame loading, API timeouts, and race conditions, particularly when using limited GPU resources. Compatibility issues between LangChain's async processes and Django's sync framework may also lead to incomplete responses. Running on a GPU-less VM further restricts performance for these tasks.

To address these challenges, future work should focus on: (1) implementing asynchronous backend processing with Django Channels or Celery to improve responsiveness; (2) caching intermediate outputs to avoid redundant computations; (3) offloading visualization displaying to the client side using lightweight JavaScript libraries or WebGL; (4) adopting lazy loading or progressive generation within the iFrame to speed up user-perceived performance; and (5) deploying on GPU-enabled VMs to accelerate model inference and explanation generation. Optimizing model execution through techniques like reduced precision or selective visualization will also reduce resource demands and improve scalability.

Further improvements include more interactive and customizable visualizations, and lighter-weight XAI alternatives to reduce overhead. We also plan to move from upload-based analysis to real-time log monitoring, enabling continuous anomaly detection and explanation. Lastly, expanding support for diverse log types will help broaden the system's applicability. Together, these efforts aim to deliver a faster, more scalable, and trustworthy explainable AI framework.

**Acknowledgments.** This work was supported by European Union NEURO-CLIMA(101137711); Research Council of Finland (323630); and Business Finland (8754/31/2022).